\documentclass[11pt,a4paper]{article}
\usepackage[hyperref]{acl2020}
\usepackage{times}
\usepackage{latexsym}

\usepackage{microtype}
\usepackage{amsmath}
\usepackage{amsfonts}
\usepackage{algorithm}
\usepackage{algpseudocode}
\usepackage{booktabs}
\usepackage{xparse}
\usepackage{graphicx}
\usepackage{multicol}
\usepackage{multirow}
\aclfinalcopy 
\usepackage{tikz}
\usetikzlibrary{shapes,arrows}
\usepackage{natbib}
\usepackage{caption}
\usepackage{subcaption}
\usepackage{tikz-dependency}
\usepackage{wrapfig}
\usepackage{pgfplots}
\usepackage{filecontents}
\usepackage[T1]{fontenc}
\usepackage{subfiles}
\usepackage{readarray}
\usepackage{xcolor}
\usepackage{import}
\usepackage{hhline}

\newcommand{\xvec}{\mathbf{x}}
\newcommand{\yvec}{\mathbf{y}}
\newcommand{\bvec}{\mathbf{b}}

\newcommand{\Wvec}{\mathbf{W}}

\newcommand{\rvec}{\mathbf{r}}

\newcommand{\mcL}{\mathcal{L}}
\newcommand{\mcT}{\mathcal{T}}
\newcommand{\mcY}{\mathcal{Y}}
\newcommand{\mcZ}{\mathcal{Z}}
\newcommand{\mcV}{\mathcal{V}}

\title{{S}tructure-{L}evel {K}nowledge {D}istillation {F}or\\ {M}ultilingual {S}equence {L}abeling}

\author{Xinyu Wang$^{\diamond}$, Yong Jiang$^{\dagger}$, Nguyen Bach$^{\dagger}$, Tao Wang$^{\dagger}$, Fei Huang$^{\dagger}$,  Kewei Tu$^{\diamond}$\thanks{\hspace{1mm} Kewei Tu is the corresponding author. This work was conducted when Xinyu Wang was interning at Alibaba DAMO Academy.} \\
 $^\diamond$School of Information Science and Technology, ShanghaiTech University \\
 Shanghai Engineering Research Center of Intelligent Vision and Imaging \\
 Shanghai Institute of Microsystem and Information Technology, Chinese Academy of Sciences \\
 University of Chinese Academy of Sciences \\
 $^\dagger$DAMO Academy, Alibaba Group \\
  {\tt \{wangxy1,tukw\}@shanghaitech.edu.cn} \\
  {\tt \{yongjiang.jy,nguyen.bach,leeo.wangt,f.huang\}@alibaba-inc.com} \\
 }

\date{}
\pgfplotsset{compat=1.14} 
\begin{document}
\maketitle
\begin{abstract}


Multilingual sequence labeling is a task of predicting label sequences using a single unified model for multiple languages. Compared with relying on multiple monolingual models, using a multilingual model has the benefit of a smaller model size, easier in online serving, and generalizability to low-resource languages. However, current multilingual models still underperform individual monolingual models significantly due to model capacity limitations. In this paper, we propose to reduce the gap between monolingual models and the unified multilingual model by distilling the structural knowledge of several monolingual models (teachers) to the unified multilingual model (student). We propose two novel KD methods based on structure-level information: (1) approximately minimizes the distance between the student's and the teachers' structure-level probability distributions, (2) aggregates the structure-level knowledge to local distributions and minimizes the distance between two local probability distributions. Our experiments on 4 multilingual tasks with 25 datasets show that our approaches outperform several strong baselines and have stronger zero-shot generalizability than both the baseline model and teacher models.
\end{abstract}


\section{Introduction}
Sequence labeling is an important task in natural language processing. Many tasks such as named entity recognition (NER) and part-of-speech (POS) tagging can be formulated as sequence labeling problems and these tasks can provide extra information to many downstream tasks and products such as searching engine, chat-bot and syntax parsing \cite{Jurafsky:2009:SLP:1214993}. Most of the previous work on sequence labeling focused on monolingual models, and the work on multilingual sequence labeling mainly focused on cross-lingual transfer learning to improve the performance of low-resource or zero-resource languages \cite{johnson-etal-2019-cross,huang-etal-2019-cross,rahimi-etal-2019-massively,huang-etal-2019-matters,keung-etal-2019-adversarial}, but their work still trains monolingual models. However, it would be very resource consuming considering if we train monolingual models for all the 7,000+ languages in the world. Besides, there are languages with limited labeled data that are required for training. Therefore it is beneficial to have a single unified multilingual sequence labeling model to handle multiple languages
, while less attention is paid to the unified multilingual models due to the significant difference between different languages. Recently, Multilingual BERT (M-BERT) \cite{devlin-etal-2019-bert} is surprisingly good at zero-shot cross-lingual model transfer on tasks such as NER and POS tagging \cite{pires-etal-2019-multilingual}. M-BERT bridges multiple languages and makes training a multilingual sequence labeling model with high performance possible \cite{wu-dredze-2019-beto}. However, accuracy of the multilingual model is still inferior to monolingual models that utilize different kinds of strong pretrained word representations such as contextual string embeddings (Flair) proposed by \citet{akbik-etal-2018-contextual}.

To diminish the performance gap between monolingual and multilingual models, we propose to utilize knowledge distillation to transfer the knowledge from several monolingual models with strong word representations into a single multilingual model. Knowledge distillation \cite{Bucilua:2006:MC:1150402.1150464,44873} is a technique that first trains a strong teacher model and then trains a weak student model through mimicking the output probabilities \cite{44873,NIPS2018_7980,DBLP:journals/corr/abs-1902-03393} or hidden states \cite{Romero2014FitNetsHF,GraphKD} of the teacher model. The student model can achieve an accuracy comparable to that of the teacher model and usually has a smaller model size through KD. 
Inspired by KD applied in neural machine translation (NMT) \cite{kim-rush-2016-sequence} and multilingual NMT \cite{tan2018multilingual}, our approach contains a set of monolingual teacher models, one for each language, and a single multilingual student model. Both groups of models are based on BiLSTM-CRF \cite{lample-etal-2016-neural,ma-hovy-2016-end}, one of the state-of-the-art models in sequence labeling. 
In BiLSTM-CRF, the CRF layer models the relation between neighbouring labels which leads to better results than simply predicting each label separately based on the BiLSTM outputs. However, the CRF structure models the label sequence globally with the correlations between neighboring labels, which increases the difficulty in distilling the knowledge from the teacher models. In this paper, we propose two novel KD approaches that take structure-level knowledge into consideration for multilingual sequence labeling. 
To share the structure-level knowledge, we either minimize the difference between the student's and the teachers' distribution of global sequence structure directly through an approximation approach or aggregate the global sequence structure into local posterior distributions and minimize the difference of aggregated local knowledge.
Experimental results show that our proposed approach boosts the performance of the multilingual model in 4 tasks with 25 datasets. Furthermore, our approach has better performance in zero-shot transfer compared with the baseline multilingual model and several monolingual teacher models.

\begin{figure*}[t]
\begin{center}
\includegraphics[width=\textwidth]{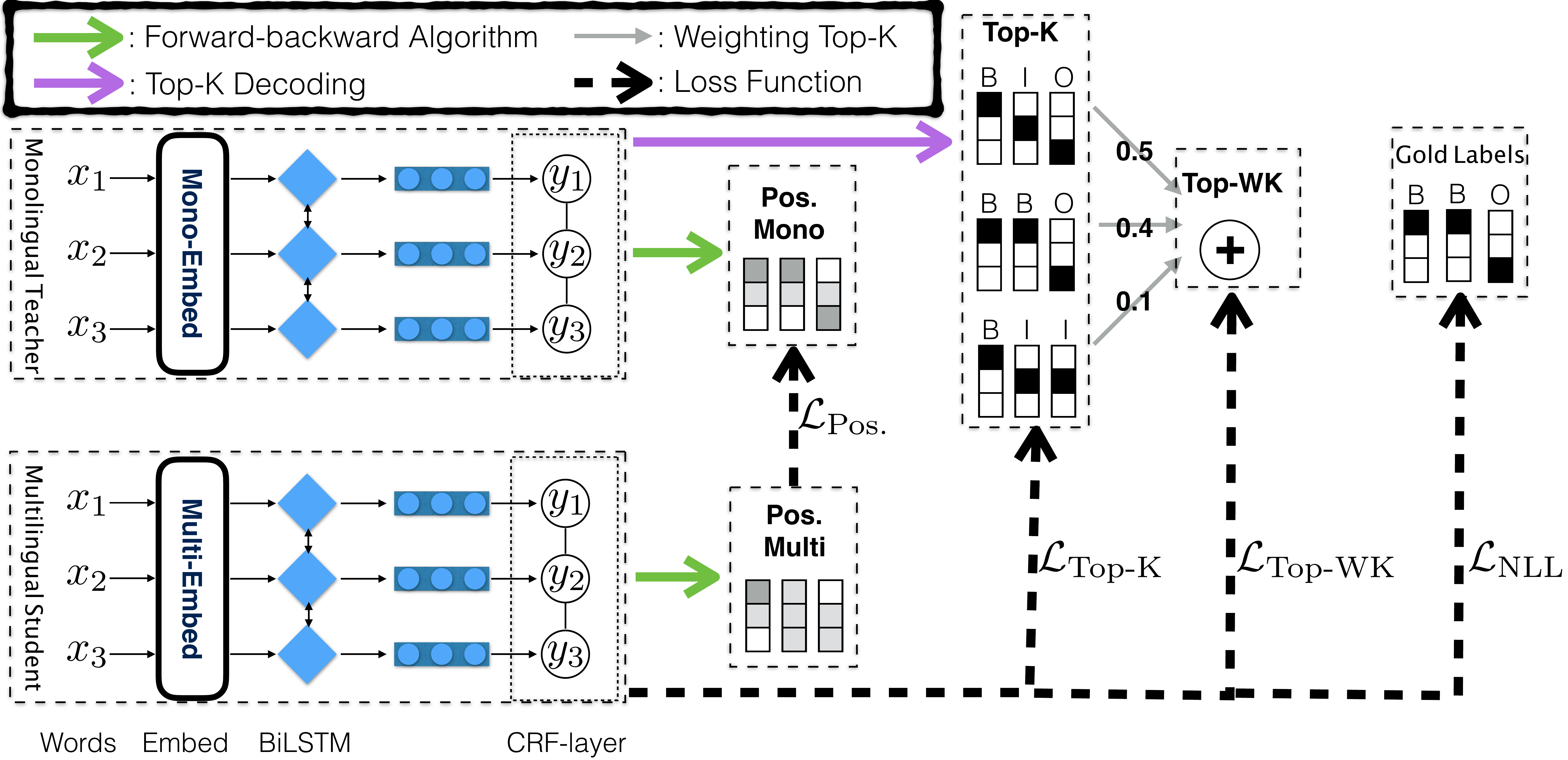}
\caption{Structure-level knowledge distillation approaches. \textit{Mono/Multi} represents Monolingual and Multilingual, respectively. \textit{Pos.} represents the posterior distribution.}
\label{fig:approach}
\end{center}
\end{figure*}
\section{Background}
\subsection{Sequence Labeling}
\label{sec:seq_lab}
BiLSTM-CRF \cite{lample-etal-2016-neural,ma-hovy-2016-end} is one of the most popular approaches to sequence labeling. Given a sequence of $n$ word tokens $\xvec = \{x_1, \cdots, x_n\}$ and the corresponding sequence of gold labels $\yvec^* = \{y_1^*, \cdots, y_n^*\}$, we first feed the token representations of $\xvec$ into a BiLSTM to get the contextual token representations $\rvec = \{\rvec_1, \cdots, \rvec_n\}$.
The conditional probability $p(\yvec|\xvec)$ is defined by:
\begin{align}
    \psi(y', y, \rvec_i) &= \exp(\Wvec_{y}^{T} \rvec_i + \bvec_{y',y}) \label{eq:psi}
\end{align}
\begin{align}
p(\yvec|\xvec) &= \frac{\prod\limits_{i=1}^{n} \psi(y_{i-1}, y_i, \rvec_i)}{\sum\limits_{\yvec' \in \mathcal{Y}(\xvec)} \prod\limits_{i=1}^{n} \psi(y'_{i-1}, y'_i, \rvec_i)}\label{eq:sentprob}
\end{align}
where $\mathcal{Y}(\xvec)$ denotes the set of all possible label sequences for $\xvec$,
$\psi$ is the potential function, $\Wvec_{y}$ and $\bvec_{y',y}$ are parameters and $y_0$ is defined to be a special start symbol. $\Wvec_{y}^{T}r_i$ and $\bvec_{y',y}$ are usually called emission and transition scores respectively. During training, the negative log-likelihood loss for an input sequence is defined by:
\begin{displaymath}
\mcL_{\text{NLL}} = - \log p(\yvec^*|\xvec)
\end{displaymath}

BiLSTM-Softmax approach to sequence labeling reduces the task to a set of label classification problem by disregarding label transitions and simply feeding the emission scores $\Wvec^{T}\mathbf{r}_i$ into a softmax layer to get the probability distribution of each variable $y_i$.
\begin{align}
p(y_i|\xvec) = \mathrm{softmax}(\Wvec^{T}\mathbf{r}_i) \label{eq:softmax}
\end{align}
The loss function then becomes:
\begin{displaymath}
\mcL_{\text{NLL}} = - \sum\limits_{i=1}^{n} \log p(y^*_{i}|\xvec)
\end{displaymath}
In spite of its simplicity, this approach ignores correlations between neighboring labels and hence does not adequately model the sequence structure. Consequently, it empirically underperforms the first approach in many applications. 

\subsection{Knowledge Distillation}
A typical approach to KD is training a student network by imitating a teacher's predictions \cite{44873}. The simplest approach to KD on BiLSTM-Softmax sequence labeling follows Eq. \ref{eq:softmax} and performs \textbf{token-level} distillation through minimizing the cross-entropy loss between the individual label distributions predicted by the teacher model and the student model:
\begin{align}
&\mcL_{\text{Token}} = \nonumber\\
&- \sum\limits_{i=1}^{n} \sum\limits_{j=1}^{|\mcV|} p_t(y_{i}=j|\xvec) \log p_s(y_{i}=j|\xvec)  \label{eq:softmax-kd}
\end{align}
where $p_t(y_{i}=j|\mathbf{x})$ and $p_s(y_{i}=j|\mathbf{x})$ are the label distributions predicted by the teacher model and the student model respectively and $|\mcV|$ is the number of possible labels. The final loss of the student model combines the KD loss and the negative log-likelihood loss:
\begin{displaymath}
L = \lambda \mcL_{\text{Token}} + (1-\lambda) \mcL_{\text{NLL}}
\end{displaymath}
where $\lambda$ is a hyperparameter. As pointed out in Section \ref{sec:seq_lab}, however, sequence labeling based on Eq. \ref{eq:softmax} has the problem of ignoring structure-level knowledge.
In the BiLSTM-CRF approach, we can also apply an \textbf{Emission} distillation through feeding emission scores in Eq. \ref{eq:softmax} and get emission probabilities $\tilde{p}(y_i|\xvec)$, then the loss function becomes:
\begin{align}
&\mcL_{\text{Emission}} = \nonumber\\
& - \sum\limits_{i=1}^{n} \sum\limits_{j=1}^{|\mcV|} \tilde{p}_t(y_{i}=j|\xvec) \log \tilde{p}_s(y_{i}=j|\xvec) \label{eq:token-kd}
\end{align}

\begin{algorithm}[t]
\small
\caption{KD for Multilingual Sequence Labeling}\label{alg:1}
\begin{algorithmic}[1]
\State \textbf{Input}: Training corpora $\mathcal{D}=\{D^1,\dots,D^l\}$ with $l$ languages, monolingual models $\mcT=\{T^1,\dots,T^l\}$ pretrained on the corresponding training corpus, learning rate $\eta$, multilingual student model $\mathcal{M}$ with parameters $\theta$, total training epochs $\mathcal{S}$, loss interpolation coefficient $\lambda$, interpolation annealing rate $\tau$. 
\State \textbf{Initialize}: Randomly initialize multilingual model parameters $\theta$. Set the current training epoch $S$ = $0$, current loss interpolation $\lambda$ = $1$. Create an new empty training dataset $\hat{\mathcal{D}}$.
\State 
\State \textbf{for} $D^i\in \mathcal{D}$ \textbf{do}
\State ~~~~\textbf{for} $(\xvec^i_j,\yvec^i_j) \in D^i$ \textbf{do}
\State ~~~~~~~~ Teacher model $T_i$ reads the input $\xvec^i_j$ and predicts probability distributions $\hat{p}^i_j$ required for KD.
\State ~~~~~~~~ Append $(\xvec^i_j,\yvec^i_j,\hat{p}^i_j)$ into the new training dataset $\hat{\mathcal{D}}$.
\State ~~~~\textbf{end for}
\State \textbf{end for}
\State 
\State \textbf{while} $S < \mathcal{S}$ \textbf{do}
\State ~~~~ $S$ = $S+1$.
\State ~~~~ \textbf{for} mini-batch $(\xvec,\yvec,\hat{p})$ sampled from $\hat{\mathcal{D}}$ \textbf{do}
\State ~~~~~~~~ Compute the KD loss $\mcL_{\text{KD}}(\xvec,\hat{p})$.
\State ~~~~~~~~ Compute the golden target loss $\mcL_{\text{NLL}}(\xvec,\yvec)$.
\State ~~~~~~~~ Compute the final loss $L=\lambda \mcL_{\text{KD}}+(1-\lambda)\mcL_{\text{NLL}}$.
\State ~~~~~~~~ Update $\theta$: $\theta$ = $\theta$ - $\eta * \partial \mathcal{L} / \partial \theta$ .
\State ~~~~ \textbf{if} $\lambda - \tau > 0$ \textbf{do}
\State ~~~~~~~~ Update interpolation factor $\lambda$: $\lambda = \lambda - \tau$ 
\State ~~~~ \textbf{else} 
\State ~~~~~~~~ Update interpolation factor $\lambda$: $\lambda = 0$
\State ~~~~ \textbf{end if}
\State \textbf{end while}
\end{algorithmic}
\end{algorithm}

\section{Approach}
In this section, we propose two approaches to learning a single multilingual sequence labeling model (student) by distilling structure-level knowledge from multiple mono-lingual models. The first approach approximately minimizes the difference between structure-level probability distributions predicted by the student and teachers. The second aggregates structure-level knowledge into local posterior distributions and then minimizes the difference between local distributions produced by the student and teachers. Our approaches are illustrated in Figure \ref{fig:approach}.

Both the student and the teachers are BiLSTM-CRF models \cite{lample-etal-2016-neural,ma-hovy-2016-end}, one of the state-of-the-art models in sequence labeling. A BiLSTM-CRF predicts the distribution of the whole label sequence structure, so token-level distillation is no longer possible and structure-level distillation is required.

\subsection{Top-K Distillation}
\label{sec:Top-K}
Inspired by \citet{kim-rush-2016-sequence}, 
we propose to encourage the student to mimic the teachers' global structural probability distribution over all possible label sequences:
\begin{align}
\mcL_{\text{Str}} = - \sum\limits_{\yvec\in \mcY(\xvec)} p_t(\yvec|\xvec) \log p_s(\yvec|\xvec) \label{eq:strkd}
\end{align}
However, $|\mcY(\xvec)|$ is exponentially large as it represents all possible label sequences. 
We propose two methods to alleviates this issue through efficient approximations of $p_t(\yvec|\xvec)$ using the $k$-best label sequences.
\begin{table*}[t!]
\small
\centering
\begin{tabular}{c|c|c||cccc||c|ccc|c}
\hline\hline
 \multicolumn{3}{c||}{\multirow{2}{*}{$\psi(y_{k-1}, y_k, \rvec_k)$}} & \multicolumn{4}{c||}{\bf \textsc{Label Seq.
Probs.}} & \multicolumn{5}{c}{\bf\textsc{Structural Knowledge}}\\
\hhline{~~~||----||-|---|-}
\multicolumn{3}{c||}{} & $y_1$            & $y_2$ & $y_3$ & \bf Prob. &                        & $y_1$    & $y_2$   & $y_3$    & \bf Weights \\
\hline
\hline
 \multicolumn{3}{c||}{$k=2$}& $F$              & $F$  & $F$  & 0.035  & \multirow{2}{*}{\bf Top-2}                  & $T$     & $T$    & $F$     & 0.57    \\
\hhline{---||~~~~||~|~~~|~}
$y_{k-1}$\texttt{\symbol{92}} $y_{k}$      & $y_2=F$ & $y_2=T$   & $F$              & $F$  & $T$  & 0.316  &                        & $F$     & $F$    & $T$     & 0.43    \\
\hhline{---||~~~~||-|---|-}
$y_1=F$ & 2  & 1/2  & $F$              & $T$  & $F$  & 0.105  & $\alpha(y_k=F)$          & 1.00  & 2.50 & 10.83 &         \\
\hhline{---||~~~~||~|~~~|~}
$y_1=T$ & 1/2  & 2 & $F$              & $T$  & $T$  & 0.007  & $\alpha(y_k=T)$          & 1.00  & 2.50 & 8.13  &         \\
\hhline{---||~~~~||-|---|~}
\hhline{---||~~~~||~|~~~|~}
 \multicolumn{3}{c||}{$k=3$} & $T$              & $F$  & $F$  & 0.009  & $\beta(y_k=F)$           & 8.79  & 3.33 & 1.00  &         \\
\hhline{---||~~~~||~|~~~|~}
  $y_{k-1}$\texttt{\symbol{92}} $y_{k}$    & $y_3=F$ & $y_3=T$  & $T$              & $F$  & $T$  & 0.079  & $\beta(y_k=T)$           & 10.17 & 4.25 & 1.00  &         \\
\hhline{---||~~~~||-|---|~}
$y_2=F$ & 1/3  & 3  & $T$              & $T$  & $F$  & 0.422  & $q(y_k=F|\mathbf{x})$    & 0.46  & 0.44 & 0.57  &         \\
\hhline{---||~~~~||~|~~~|~}
$y_2=T$ & 4  & 1/4 & $T$              & $T$  & $T$  & 0.026  & $q(y_k=T|\mathbf{x})$    & 0.54  & 0.56 & 0.43  &   \\
\hline
\hline
\end{tabular}
\caption{Example of computing the structural knowledge for a sequence of 3 tokens with a label set of $\{T,F\}$. $\psi(y_{k-1}, y_k, \rvec_k)$ represents the potential formulated in Eq. \ref{eq:psi}. Each \textsc{\bf Label Seq. Probs.} is defined in Eq. \ref{eq:sentprob} for the corresponding label sequence. \textbf{Top-2} represents the two label sequences with the highest scores and \textbf{Weights} are their corresponding weights for KD (Eq. \ref{eq:kbest}, \ref{eq:weighted}). $\alpha(y_k)$, $\beta(y_k)$ and the posterior distribution $q(y_k|\mathbf{x})$ are computed based on Eq. \ref{eq:alpha}, \ref{eq:beta} and \ref{eq:posterior} respectively. We assume that $\psi(y_0, y_1, \rvec_1)=1$ regardless of whether $y_1$ is $T$ or $F$.}
\label{tab:example}
\end{table*}
\paragraph{Top-K}
Eq. \ref{eq:strkd} can be seen as computing the expected student log probability with respect to the teacher's structural distribution:
\begin{align}
\mcL_{\text{Str}} = -\mathbb{E}_{p_t(\yvec|\xvec)}[\log p_s(\yvec|\xvec)]\label{eq:expect}
\end{align}
The expectation can be approximated by sampling from the teacher's distribution $p_t(\yvec|\xvec)$. However, unbiased sampling from the distribution is difficult. We instead apply a biased approach that regards the $k$-best label sequences predicted by the teacher model as our samples. We use a modified Viterbi algorithm to predict the $k$-best label sequences $\mcT=\{\hat{\yvec}^1,\dots,\hat{\yvec}^k\}$. Eq. \ref{eq:expect} is then approximated as:
\begin{align}
\mcL_{\text{Top-K}} = - \frac{1}{k}\sum\limits_{\hat{\yvec}\in \mcT} \log p_s(\hat{\yvec}|\xvec) \label{eq:kbest}
\end{align}
This can also be seen as data augmentation through generating $k$ pseudo target label sequences for each input sentence by the teacher.

\paragraph{Weighted Top-K}
The Top-K method is highly biased in that the approximation becomes worse with a larger $k$ . A better method is to associate weights to the $k$ samples to better approximate $p_t(\yvec|\xvec)$.
\begin{align*}
p^{\prime}_t(\yvec|\xvec) & = 
\begin{cases}
\frac{p_t(\yvec|\xvec)}{\sum\limits_{\hat{\yvec}\in \mcT}p_t(\hat{\yvec}|\xvec)} & \yvec \in \mcT\\
0 & \yvec \notin \mcT \\
\end{cases}
\end{align*}
Eq. \ref{eq:expect} is then approximated as:
\begin{align}
\mcL_{\text{Top-WK}} = - \sum\limits_{\yvec\in \mcT} p^{\prime}_t(\yvec|\xvec) \log p_s(\yvec|\xvec) \label{eq:weighted}
\end{align}
This can be seen as the student learning weighted pseudo target label sequences produced by the teacher for each input sentence.

The \textbf{Top-K} approach is related to the previous work on model compression in neural machine translation \cite{kim-rush-2016-sequence} and multilingual neural machine translation \cite{tan2018multilingual}.
In neural machine translation, producing $k$-best label sequences is intractable in general and in practice, beam search decoding has been used to approximate the $k$-best label sequences. However, for linear-chain CRF model, $k$-best label sequences can be produced exactly with the modified Viterbi algorithm.

\subsection{Posterior Distillation}
\label{sec:posterior}
The \textbf{Top-K} is approximate with respect to the teacher's structural distribution and still is slow on large $k$. 
Our second approach tries to distill structure-level knowledge based on tractable local (token-wise) distributions $q(y_k|\xvec)$, which can be exactly computed.
\begin{align}
q(y_k|\xvec)&=\sum\limits_{\{y_1,\dots,y_n\}\setminus y_k} p(y_1,\dots,y_n|\xvec) \nonumber\\
&=\frac{\sum\limits_{\{y_1,\dots,y_n\}\setminus y_k} \prod\limits_{i=1}^{n} \psi(y_{i-1}, y_i, \rvec_i)}{\mcZ} \label{eq:posterior}\\
&\propto \alpha(y_k) \times \beta(y_k) \nonumber \\ 
\alpha(y_k)&=\sum\limits_{\{y_0,\dots,y_{k-1}\}} \prod\limits_{i=1}^{k} \psi(y_{i-1}, y_i, \rvec_i) \label{eq:alpha}\\
\beta(y_k) &= \sum\limits_{\{y_{k+1},\dots,y_n\}} \prod\limits_{i=k+1}^{n} \psi(y_{i-1}, y_i, \rvec_i)\label{eq:beta}
\end{align}
where $\mcZ$ is the denominator of Eq. \ref{eq:sentprob} that is usually called the \textit{partition function} and $\alpha(y_k)$ and $\beta(y_k)$ are calculated in forward and backward pass utilizing the forward-backward algorithm. We assume that $\beta(y_n)=1$.

Given the local probability distribution for each token, we define the KD loss function in a similar manner with the token-level distillation in Eq. \ref{eq:token-kd}. 

\begin{align}
\mcL_{\text{Pos.}} = - \sum\limits_{i=1}^{n} \sum\limits_{j=1}^{|\mcV|} q_t(y_{i}=j|\xvec) \log q_s(y_{i}=j|\xvec) \label{eq:pos-kd}
\end{align}

The difference between token-level distillation and posterior distillation is that posterior distillation is based on BiLSTM-CRF and conveys global structural knowledge in the local probability distribution. 

Posterior distillation has not been used in the related research of knowledge distillation in neural machine translation because of intractable computation of local distributions. In sequence labeling, however, local distributions in a BiLSTM-CRF can be computed exactly using the forward-backward algorithm. 

An example of computing the structural knowledge discussed in this and last subsections is shown in Table \ref{tab:example}.


\subsection{Multilingual Knowledge Distillation}
Let $\mathcal{D}=\{D^1,\dots,D^l\}$ denotes a set of training data with $l$ languages. $D^i$ denotes the corpus of the $i$-th language that contains multiple sentence and label sequence pairs $D^i=\{(\xvec^i_j,\yvec^i_j)\}_{j=1}^{m_i}$. To train a single multilingual student model from multiple monolingual pretrained teachers, for each input sentence, we first use the teacher model of the corresponding language to predict the pseudo targets ($k$-best label sequences or posterior distribution for posterior distillation). Then the student jointly learns from the gold targets and pseudo targets in training by optimizing the following loss function:
\begin{align*}
\mcL_{\text{ALL}} = \lambda \mcL_{\text{KD}} + (1-\lambda) \mcL_{\text{NLL}}
\end{align*}
where $\lambda$ decreases from 1 to 0 throughout training following \citet{clark-etal-2019-bam}, $\mcL_{\text{KD}}$ is one of the Eq. \ref{eq:token-kd}, \ref{eq:kbest}, \ref{eq:weighted}, \ref{eq:pos-kd} or an averaging of Eq. \ref{eq:weighted}, \ref{eq:pos-kd}. The overall distillation process is summarized in Algorithm \ref{alg:1}.  

\begin{table*}[t!]
\setlength\tabcolsep{2pt}
\centering
\small
\begin{tabular}{c|l|cccc|c||ccccc|c}
\hline\hline
& \bf Task &  \multicolumn{5}{c||}{\bf CoNLL NER} &  \multicolumn{6}{c}{\bf SemEval 2016 Aspect Extraction}\\
\hline
&\bf Approach &   \bf English & \bf Dutch    & \bf Spanish    & \bf German    & \bf Avg.  & \bf Turkish    & \bf Spanish    & \bf Dutch    & \bf English    & \bf Russian    & \bf Avg.  \\
\hline
\multirow{3}{*}{\textsc{Ref}}&\bf \textsc{Teachers}               & 92.43 & 91.90 & 89.19 & 84.00 & 89.38 & 59.29 & 74.29 & 72.85 & 72.80 & 71.77 & 70.20 \\
&\bf \textsc{Softmax} & 90.08 & 88.99 &  87.72 & 81.40 & 87.05 & 52.39 & 71.54 & 68.86 & 65.87 & 66.85 & 65.10 \\
& \bf \textsc{Token} & 90.02 & 88.87 & 88.24 & 81.30 & 87.11 & 52.56 & 72.12 & 69.33 & 66.81 & 67.20 & 65.61 \\
\hline
\multirow{2}{*}{\textsc{Base}}&\bf \textsc{Baseline}               & 90.13 & 89.11 & 88.06 & 82.16 & 87.36 & 55.79 & 72.02 & 69.35 & 67.54 & 68.02 & 66.54 \\
& \bf \textsc{Emission}            & 90.28 & 89.31 & 88.65 & 81.96 & 87.55 & 51.52 & 72.60 & 69.10 & 67.21 & 68.52 & 65.79 \\
\hline
\multirow{4}{*}{\textsc{Ours}}&\bf \textsc{Top-K}                   & 90.57 &  89.33  & 88.61&   81.99 &  87.62  & 55.74 & 73.13 & 69.81 & 67.99 & 69.21 & 67.18 \\
&\bf \textsc{Top-WK}      & 90.52&   89.24&   88.64&   82.15&   87.64  & 56.40 & 72.81 & 69.33 & \textbf{68.16} & 69.42 & 67.22 \\
&\bf \textsc{Posterior}              & \textbf{90.68} & 89.41 & 88.57 & 82.22 & 87.72 & \textbf{56.69} & 73.47 & 69.98 & 68.11 & 69.22 & \textbf{67.49} \\
&\bf \textsc{Pos.+Top-WK} & 90.53 & \textbf{89.58} & \textbf{88.66} & \textbf{82.31} & \textbf{87.77} & 55.00 & \textbf{73.97} & \textbf{70.15} & 67.83 & \textbf{69.76} & 67.34 \\
\hline\hline
\end{tabular}
\caption{Results in F1 score of CoNLL 2002/2003 NER task and Aspect Extraction of SemEval 2016 Task 5.}
\label{tab:ner}
\end{table*}

\section{Experiment}
\subsection{Setup}
\paragraph{Dataset} We use datasets from 4 sequence labeling tasks in our experiment. 
\begin{itemize}
    \item {\bf CoNLL} NER: We collect the corpora of 4 languages from the CoNLL 2002 and 2003 shared task \cite{tjong-kim-sang-2002-introduction,tjong-kim-sang-de-meulder-2003-introduction}
    \item {\bf WikiAnn} NER \cite{pan-etal-2017-cross}: The dataset contains silver standard NER tags that are annotated automatically on 282 languages that exist in Wikipedia. We select the data of 8 languages from different language families or from different language subgroups of Indo-European languages. We randomly choose 5000 sentences from the dataset for each language except English, and choose 10000 sentences for English to reflect the abundance of English corpora in practice. We split the dataset by 8:1:1 for training/development/test.
    \item {\bf Universal Dependencies} (UD) \cite{nivre-etal-2016-universal}: We use universal POS tagging annotations in the UD datasets. We choose 8 languages from different language families or language subgroups and one dataset for each language.
    \item {\bf Aspect Extraction}: The dataset is from an aspect-based sentiment analysis task in SemEval-2016 Task 5 \cite{pontiki-etal-2016-semeval}. We choose subtask 1 of the restaurants domain which has the most languages in all domains\footnote{Subtask 1 of the restaurants domain contains 6 languages but we failed to get the French dataset as the dataset is not accessible from the provided crawling toolkit.}, and split 10\% of the training data as the development data.
\end{itemize}


\begin{table*}[t!]
\centering
\setlength\tabcolsep{2.5pt}
\small
\begin{tabular}{c|l|cccccccc|c}
\hline\hline
& \bf Approach & \bf English    & \bf Tamil    & \bf Basque    & \bf Hebrew    & \bf Indonesian    & \bf Persian    & \bf Slovenian    & \bf French    & \bf Avg.   \\
\hline
\multirow{3}{*}{\textsc{Ref}}& \bf \textsc{Teachers}           & 83.80 & 86.72 & 94.68 & 83.72 & 90.48 & 90.37 & 91.66 & 90.29 & 88.97 \\
&\bf \textsc{Softmax} & 81.86 & 80.72 & 93.72 & 77.11 & 90.64 & 90.03 & 91.05 & 88.18 & 86.66 \\
& \bf \textsc{Token} & 81.33 & 80.88 & 93.56 & 77.47 & 90.50 & 89.83 & 91.08 & 87.93 & 86.57 \\
\hline
\multirow{2}{*}{\textsc{Base}}& \bf \textsc{Baseline}               & 82.56 & 82.39 & 94.13 & 78.89 & 91.11 & 90.23 & 91.62 & 88.92 & 87.48 \\
& \bf \textsc{Emission}            & 82.54 & 82.23 & 94.37 & 78.45 & 90.92 & 89.92 & 91.56 & 89.47 & 87.43 \\
\hline
\multirow{4}{*}{\textsc{Ours}}& \bf \textsc{Top-K}                   & 82.39 & 82.94 & 94.13 & \textbf{78.93} & 90.93 & 90.12 & 91.56 & 89.25 & 87.53 \\
& \bf \textsc{Top-WK}      & 82.55 & 82.71 & 94.44 & 78.79 & 91.18 & 90.22 & 91.37 & 89.32 & 87.57 \\
& \bf \textsc{Posterior}              & \textbf{83.03} & \textbf{83.02} & 94.35 & 78.77 & \textbf{91.75} & 90.11 & \textbf{91.95} & \textbf{89.65} & \textbf{87.83} \\
& \bf Pos.+Top-WK & 82.77 & 82.81 & \textbf{94.47} & 78.87 & 91.18 & \textbf{90.31} & 91.84 & 89.42 & 87.71 \\
\hline\hline
\end{tabular}
\caption{F1 scores in the WikiAnn NER task.}
\label{tab:wikiann}
\end{table*}

\begin{table*}[t!]
\centering
\setlength\tabcolsep{2.5pt}
\small
\begin{tabular}{c|l|cccccccc|c}
\hline\hline
 & \bf Approach & \bf English    & \bf Hebrew    & \bf Japanese    & \bf Slovenian  & \bf French    & \bf Indonesian    & \bf Persian    & \bf Tamil    & \bf Avg.  \\
\hline
\multirow{3}{*}{\textsc{Ref}} & \bf \textsc{Teachers}           & 96.94 & 97.54 & 96.81 & 95.01 & 99.10 & 94.02 & 98.07 & 93.01 & 96.31 \\
&\bf \textsc{\textsc{Softmax}} & 95.61 & 96.25 & 96.59 & 90.66 & 97.94 & 92.56 & 96.62 & 86.58 & 94.10 \\
& \bf \textsc{Token} & 95.66 & 96.28 & 96.47 & 90.82 & 97.95 & 92.70 & 96.58 & 86.41 & 94.11 \\
\hline
\multirow{2}{*}{\textsc{Base}}&\bf \textsc{Baseline}               & 95.71 & 96.18 & \textbf{96.60} & 90.64 & 97.89 & 92.62 & 96.63 & 86.19 & 94.06 \\
& \bf \textsc{Emission}            & 95.63 & 96.21 & 96.52 & 90.76 & 97.98 & 92.64 & 96.61 & 86.66 & 94.13 \\
\hline
\multirow{4}{*}{\textsc{Ours}}&\bf \textsc{Top-K}                   & \textbf{95.74} & 96.27 & 96.56 & 90.66 & 97.96 & 92.58 & 96.64 & 86.57 & 94.12 \\
& \bf \textsc{Top-WK}      & 95.68 & 96.23 & 96.58 & 90.73 & 97.89 & 92.62 & 96.62 & 86.74 & 94.14 \\
& \bf \textsc{Posterior}              & 95.71 & \textbf{96.34} & 96.59 & \textbf{90.91} & 97.99 & \textbf{92.72} & 96.69 & \textbf{87.36} & \textbf{94.29} \\
& \bf \textsc{Pos.+Top-WK} & \textbf{95.74} & 96.27 & 96.47 & 90.84 & \textbf{98.02} & 92.58 & \textbf{96.73} & 86.97  & 94.20 \\
\hline\hline
\end{tabular}
\caption{Accuracies in UD POS tagging.}
\label{tab:upos}
\end{table*}

\paragraph{Model Configurations} 
In our experiment, all the word embeddings are fixed
and M-BERT token embeddings are obtained by average pooling. We feed the token embeddings into the BiLSTM-CRF for decoding. The hidden size of the BiLSTM layer is 256 for the monolingual teacher models and 600 or 800 for the multilingual student model depending on the dataset as larger hidden size for the multilingual model results in better performance in our experiment. The settings of teacher and student models are as follows:
\begin{itemize}
    \item {\bf Monolingual Teachers:} Each teacher is trained with a dataset of a specific language. We use M-BERT concatenated with language-specific \textit{Flair} \cite{akbik-etal-2018-contextual} embeddings and \textit{fastText} \cite{bojanowski2017enriching} word embeddings as token embeddings\footnote{We use fastText + M-BERT instead if the Flair embedding is not available for a certain language.} for all the monolingual teacher models. 
    \item {\bf Multilingual Student:} The student model is trained with the datasets of all the languages combined. We only use M-BERT as token embeddings for the multilingual student model. 
\end{itemize}


\paragraph{Training} For model training, the mini-batch size is set to 2000 tokens. We train all models with SGD optimizer with a learning rate of 0.1 and anneal the learning rate by 0.5 if there is no improvements on the development set for 10 epochs. For all models, we use a single NVIDIA Tesla V100 GPU for training including the student model. We tune the loss interpolation anneal rate in $\{0.5,1.0\}$ and the $k$ value of Top-K ranging from $[1,10]$.

\begin{table*}[ht]
\centering
\small
\begin{tabular}{l|ccccccc|c}
\hline\hline
 & \bf Tamil & \bf Basque    & \bf Hebrew    & \bf Indonesian & \bf Persian   & \bf Slovenian    & \bf French      & \bf Avg.   \\
\hline
\bf \textsc{Teachers}&          24.98&  40.51&  25.39&  35.54&  11.05&  59.95&  60.54&  36.85  \\
\bf \textsc{Baseline}         & 37.83 & \textbf{47.80} & 47.96 & 38.71 & 16.23 & 61.22 & 59.34 & 44.15 \\
\bf \textsc{Emission}        & 37.99 & 46.69 & 47.34 & 38.52 & 16.11 & 60.75 & 59.81 & 43.89 \\
\bf \textsc{Posterior}        & \textbf{38.93} & 47.52 & 48.33 & 38.76 & \textbf{16.69} & \textbf{62.04} & \textbf{60.77} & \textbf{44.72} \\
\bf \textsc{Posterior+Top-WK} & 38.23 & 47.49 & \textbf{48.79} & \textbf{39.32} & 16.19 & 62.03 & 60.34 & 44.63 \\
\hline\hline
\end{tabular}
\caption{Results of zero-shot transfer in the NER task (CoNLL $\Rightarrow$ WikiAnn).}
\label{tab:zeroshot}
\end{table*}

\subsection{Results}
\label{sec:result}
We report results of the following approaches.
\begin{itemize}
    \item {\bf Baseline} represents training the multilingual model with the datasets of all the languages combined and without knowledge distillation.
    \item {\bf Emission} is the KD method based on Eq. \ref{eq:token-kd}.
    \item {\bf Top-K}, {\bf Top-WK} and {\bf Posterior} are our KD methods formulated by Eq. \ref{eq:kbest}, Eq. \ref{eq:weighted} and Eq. \ref{eq:pos-kd} resprectively. 
    \item {\bf Pos.+Top-WK} is a mixture of posterior and weighted Top-K distillation.
\end{itemize}
We also report the results of monolingual models as {\bf Teachers} and multilingual BiLSTM-Softmax model with token-level KD based on Eq. \ref{eq:softmax-kd} as {\bf Softmax} and {\bf Token} for reference.

Table \ref{tab:ner}, \ref{tab:wikiann}, and \ref{tab:upos} show the effectiveness of our approach on 4 tasks over 25 datasets. In all the tables, we report scores averaged over 5 runs. 

\noindent\textbf{Observation \#0. BiLSTM-Softmax models perform inferior to BiLSTM-CRF models in most cases in the multilingual setting:}
The results show that the BiLSTM-CRF approach is stronger than the BiLSTM-Softmax approach on three of the four tasks, which are consistent with previous work on sequence labeling \cite{ma-hovy-2016-end,reimers2017optimal,yang-etal-2018-design}. The token-level KD approach performs almost the same as the BiLSTM-Softmax baseline in most of the tasks except the Aspect Extraction task.

\noindent\textbf{Observation \#1. Monolingual teacher models outperform multilingual student models:}
This is probably because the monolingual teacher models are based on both multilingual embeddings M-BERT and strong monolingual embeddings (Flair/fastText). The monolingual embedding may provide additional information that is not available to the multilingual student models.
Furthermore, note that the learning problem faced by a multilingual student model is much more difficult than that of a teacher model because a student model has to handle all the languages using roughly the same model size as a teacher model.

\noindent\textbf{Observation \#2. Emission fails to transfer knowledge:}
\textbf{Emission} outperforms the baseline only on 12 out of 25 datasets. This shows that simply following the standard approach of knowledge distillation from emission scores is not sufficient for the BiLSTM-CRF models.

\noindent\textbf{Observation \#3. Top-K and Top-WK outperform the baseline:}
\textbf{Top-K} outperforms the baseline on 15 datasets. It outperforms \textbf{Emission} on average on Wikiann NER and Aspect Extraction and is competitive with \textbf{Emission} in the other two tasks. \textbf{Top-WK} outperforms the baseline on 18 datasets and it outperforms \textbf{Top-K} in all the tasks.
\noindent\textbf{Observation \#4. Posterior achieves the best performance on most of the tasks:}
The \textbf{Posterior} approach outperforms the baseline on 21 datasets and only underperforms the baseline by 0.12 on 2 languages in WikiAnn and by 0.01 on one language in UD POS tagging. It outperforms the other methods on average in all the tasks except that is slightly underperforms \textbf{Pos.+Top-WK} in the CoNLL NER task.

\noindent\textbf{Observation \#5. Top-WK+Posterior stays in between:}
\textbf{Pos.+Top-WK} outperforms both \textbf{Top-WK} and \textbf{Posterior} only in the CoNLL NER task. In the other three tasks, its performance is above that of \textbf{Top-WK} but below that of \textbf{Posterior}.


\begin{table}[t]
\centering
\small
\begin{tabular}{l|cc}
\hline\hline
& \textbf{NER} & \textbf{POS} \\
\hline
{\bf\textsc{Teachers}} & 41.85 & 56.01 \\
{\bf\textsc{Baseline}} & 50.86 & 84.11 \\
{\bf\textsc{Emission}} & 50.19 & 84.17 \\
{\bf\textsc{Posterior}} & \textbf{51.43} & \textbf{84.28} \\
{\bf\textsc{Posterior+Top-K}} & 51.14 & 84.24\\
\hline\hline
\end{tabular}
\caption{Averaged results of zero-shot transfer on another 28 languages of the NER task and 24 languages of the POS tagging task.}
\label{tab:averaged_zs}
\end{table}

\subsection{Zero-shot Transfer}
We use the monolingual teacher models, multilingual baseline models and our \textbf{Posterior} and \textbf{Pos.+Top-WK} models trained on the CoNLL NER datasets to predict NER tags on the test sets of 7 languages in WikiAnn that used in Section \ref{sec:result}. Table \ref{tab:zeroshot} shows the results. For the teacher models, we report the maximum score over all the teachers for each language. The results show that multilingual models significantly outperform the teacher models. For languages such as Tamil and Hebrew, which are very different from the languages in the CoNLL datasets, the performance of the teacher models drops dramatically compared with the multilingual models. It shows that the language specific features in teacher models limits their generalizability on new languages. Our multilingual models, \textbf{Posterior} and \textbf{Pos.+Top-WK} outperform the baseline on all the languages. \textbf{Emission} slightly underperforms \textit{Baseline}, once again showing its ineffectiveness in knowledge distillation. 

We also conduct experiments on zero-shot transferring over other 28 languages on WikiAnn NER datasets and 24 languages on UD POS tagging datasets. The averaged results are shown in Table \ref{tab:averaged_zs}. The NER experiment shows that our approaches outperforms \textbf{Baseline} on 24 out of 28 languages and the \textbf{Posterior} is stronger than \textbf{Pos.+Top-WK} by 0.29 F1 score on average. The POS tagging experiment shows that our approach outperforms \textbf{Baseline} on 20 out of 24 languages. For more details, please refer to the Appendices \ref{sec:supplemental}. 

\subsection{KD with Weaker Teachers}

\begin{table}[t]
\setlength\tabcolsep{2.5pt}
\centering
\small
\begin{tabular}{c|cccc|c}
\hline\hline
 &\bf English & \bf Dutch    & \bf Spanish    & \bf German    & \bf Avg.\\
\hline
\bf\textsc{Teachers} & 90.63 & 89.65 & 88.05 & 81.81 & 87.54 \\
\hline
\bf\textsc{Baseline}      & 90.13 & 89.11 & 88.06 & \textbf{82.16} & 87.36 \\
\bf\textsc{Posterior} & \textbf{90.57} & \textbf{89.17} & \textbf{88.61} & \textbf{82.16} & \textbf{87.63} \\
\hline\hline
\end{tabular}
\caption{Posterior distillation with weaker teachers.}
\label{tab:M-BERT_posterior}
\end{table}

To show the effectiveness of our approach, we train weaker monolingual teachers using only M-BERT embeddings on four datasets of the \textbf{CoNLL} NER task. We run \textbf{Posterior} distillation and keep the setting of the student model unchanged. In this setting, \textbf{Posterior} not only outperforms the baseline, but also outperforms the teacher model on average. This shows that our approaches still work when the teachers have the same token embeddings as the student. By comparing Table \ref{tab:M-BERT_posterior} and \ref{tab:ner}, we can also see that stronger teachers lead to better students.

\subsection{$k$ Value in Top-K}

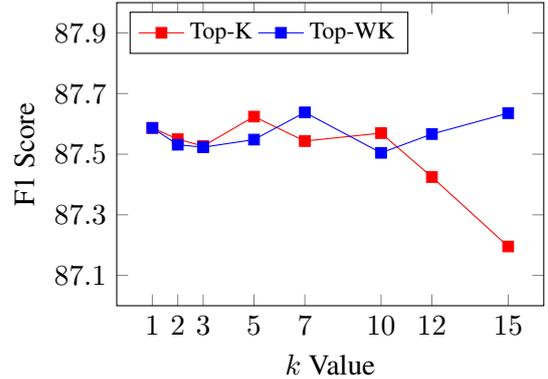
\begin{figure}[t]
\centering
\begin{tikzpicture}
    \begin{axis}[
        width=0.45\textwidth,
        height=0.35\textwidth,
        xlabel=$k$ Value,
        ylabel=F1 Score,
        legend columns=2, 
        legend pos=north west,
        legend style={font=\small},
        xtick={1,2,3,5,7,10,12,15},
        ytick={86.5,86.7,86.9,87.1,87.3,87.5,87.7,87.9,88.1},
        ymin=87.0,
        ymax=88.0,
        ]
        \addplot[red,mark=square*] table[x=kvalue,y=score] {kbest.dat};
        \addplot[blue,mark=square*] table[x=kvalue,y=score] {kbestatt.dat};
        \legend{Top-K,Top-WK}
    \end{axis}
\end{tikzpicture}
\caption{Averaged F1 scores on the CoNLL NER task versus the $k$ values of Top-K distillation. }
\label{fig:kbest}
\end{figure}

To show how the $k$ value affects the performance of \textbf{Top-K} and \textbf{Top-WK} distillation methods, we compare the models with two distillation methods and different $k$ values on the CoNLL NER task. Figure \ref{fig:kbest} shows that \textbf{Top-K} drops dramatically when $k$ gets larger while \textbf{Top-WK} performs stably. Therefore \textbf{Top-WK} is less sensitive to the hyper-parameter $k$ and might be practical in real applications.

\begin{table}[t]
\centering
\small
\begin{tabular}{l|cc}
\hline\hline
& \textbf{Training Time (hours)}\\
\hline
{\bf\textsc{Baseline}} & 11  \\
{\bf\textsc{Emission}} & 11.5  \\
{\bf\textsc{Top-WK}} & 18  \\
{\bf\textsc{Posterior}} & 16 \\
\hline\hline
\end{tabular}
\caption{Training time of the Baseline and KD approaches on CoNLL NER datasets. The training time of KD approaches includes teachers predicting and student training.}
\label{tab:time-consumption}
\end{table}

\subsection{Training Time and Memory Consumption}
We compare the training time of different approaches on the CoNLL NER task and report the results in Table \ref{tab:time-consumption}. Our \textbf{Top-WK} and \textbf{Posterior} approaches take 1.45 and 1.63 times the training time of the \textbf{Baseline} approach. For the memory consumption in training, the GPU memory cost does not vary significantly for all the approaches, while the CPU memory cost for all the KD approaches is about 2 times that of the baseline model, because training models with KD requires storing predictions of the teachers in the CPU memory.

\section{Related Work}
\paragraph{Multilingual Sequence Labeling} Many important tasks such as NER and POS tagging can be reduced to a sequence labeling problem. Most of the recent work on multilingual NER \cite{tackstrom-2012-nudging,fang-etal-2017-learning,enghoff-etal-2018-low,rahimi-etal-2019-massively,johnson-etal-2019-cross} and POS tagging \cite{snyder-etal-2009-adding,plank-agic-2018-distant} focuses on transferring the knowledge of a specific language to another (low-resource) language. For example, \citet{johnson-etal-2019-cross} proposed cross-lingual transfer learning for NER focusing on bootstrapping Japanese from English, which has a different character set than Japanese.

\paragraph{Pretrained Word Representations}
Recent progress on pretrained word representations such as ELMo \cite{peters-etal-2018-deep}, BERT \cite{devlin-etal-2019-bert} and XLNet \cite{NIPS2019_8812} significantly improve the performance of multiple NLP tasks. Multilingual BERT is a pretrained BERT model incorporating 104 languages into a single multilingual model. \citet{pires-etal-2019-multilingual} showed its ability of generalization and zero-shot transfer learning on NER and POS tagging and \citet{keung-etal-2019-adversarial} used adversarial learning with M-BERT and significantly improved zero-resource cross-lingual NER. On the tasks of NER and POS tagging, Flair embeddings \cite{akbik-etal-2018-contextual,akbik-etal-2019-pooled} is a state-of-the-art method based on character-level language models. \citet{straka2019evaluating} found that concatenating Flair embeddings with BERT embeddings outperforms other mixtures of ELMo, BERT and Flair embeddings in most of the subtasks on the CoNLL 2018 Shared Task \cite{zeman-shared-2018-conll} datasets on 54 languages, which inspired us to use M-BERT + Flair embeddings as the word representation of teachers.

\paragraph{Knowledge Distillation}
Knowledge distillation has been used to improve the performance of small models with the guidance of big models, with applications in natural language processing \cite{kim-rush-2016-sequence,kuncoro-etal-2016-distilling,tan2018multilingual,clark-etal-2019-bam,sun2019patient}, computer vision \cite{NIPS2014_5484} and speech recognition \cite{Huang2018}. For simple classification problems, there is a variety of work on tasks such as sentiment analysis \cite{clark-etal-2019-bam}, image recognition \cite{44873} and cross-lingual text classification \cite{xu-yang-2017-cross}.
For structured prediction problems, there are lines of work on neural machine translation  \cite{kim-rush-2016-sequence,tan2018multilingual}, connectionist temporal classification in the field of speech recognition \cite{Huang2018} and dependency parsing \cite{kuncoro-etal-2016-distilling,liu2018distilling}. 
Many recent researches on BERT with knowledge distillation are focused on distilling a large BERT model into a smaller one. \cite{tsai-etal-2019-small} distilled a large M-BERT model into a three layer M-BERT model for sequence labeling and achieved a competitively high accuracy with significant speed improvements. \cite{jiao2019tinybert} proposed TinyBERT for natural language understanding. \cite{sanh2019distilbert} proposed a distilled version of the BERT model which achieves a 60\% faster speed and maintains 97\% performance of the larger BERT model.


\section{Discussion on Flair/M-BERT Fine-tuning}
Previous work has discussed and empirically investigated two ways of adapting monolingual pretrained embedding models to monolingual downstream tasks \cite{peters-etal-2019-tune}: either fixing the models and using them for feature extraction, or fine-tuning them in downstream tasks. They found that both settings have comparable performance in most cases. \citet{wu-dredze-2019-beto} found that fine-tuning M-BERT with the bottom layers fixed provides further performance gains in multilingual setting. In this paper, we mainly focus on the first approach and utilize the pretrained embedding as fixed feature extractor because Flair/M-BERT finetuning is too slow for our large-scale experimental design of multilingual KD. Designing a cheap and fast fine-tuning approach for pretrained embedding models might be an interesting direction for future work. 

\section{Conclusion}
In this paper our major contributions are the two structure-level methods to distill the knowledge of monolingual models to a single multilingual model in sequence labeling: Top-K knowledge distillation and posterior distillation.
The experimental results show that our approach improves the performance of multilingual models over 4 tasks on 25 datasets. 
The analysis also shows that our model has stronger zero-shot transfer ability on unseen languages on the NER and POS tagging task. Our code is publicly available at \url{https://github.com/Alibaba-NLP/MultilangStructureKD}.
\section*{Acknowledgement}
This work was supported by the National Natural Science Foundation of China (61976139).
\bibliography{anthology,acl2020}
\bibliographystyle{acl_natbib}

\appendix
\section{Appendices}
\label{sec:supplemental}
In this appendices, we use ISO 639-1 codes\footnote{\url{https://en.wikipedia.org/wiki/List_of_ISO_639-1_codes}} to represent each language for simplification.

\subsection{Zero-shot Transfer}
Table \ref{tab:zs_ner}, \ref{tab:zs_pos} shows performance of zero-shot transfer on the NER and POS tagging datasets. Our \textbf{Posterior} approach outperforms \textbf{Baseline} in 24 out of 28 languages on NER and 20 out of 24 languages on POS tagging.

\begin{table*}[t]
\centering
\small
\begin{tabular}{l|cccccccccc}
\hline
\hline
 &  ar & be & ca & cs & da & el & eo & et & fi & gl\\
 \hline
(1): {\bf\textsc{Teacher}}  & 14.77 & 26.96 & $\mathbf{57.75}$ & 57.16 & 65.19 & 45.70 & 35.81 & 49.66 & 55.61 & 63.73 \\
(2): {\bf\textsc{Baseline}}  & 27.72 & $\mathbf{64.64}$ & 55.78 & 65.40 & 68.33 & 60.76 & 37.94 & 59.54 & 63.41 & 64.83 \\
(3): {\bf\textsc{Emission}}  & 26.92 & 63.75 & 55.30 & 64.27 & 68.09 & 59.86 & 37.28 & 59.23 & 63.68 & 64.99 \\
(4): {\bf\textsc{Posterior}}  & 27.83 & 64.62 & 56.82 & 65.69 & $\mathbf{69.08}$ & 60.66 & $\mathbf{38.44}$ & $\mathbf{60.47}$ & $\mathbf{64.03}$ & 65.07 \\
(5): {\bf\textsc{Posterior+Top-K}}  & $\mathbf{28.31}$ & 64.54 & 56.34 & 65.80 & $\mathbf{69.08}$ & $\mathbf{61.33}$ & 38.14 & 60.16 & 63.62 & $\mathbf{65.12}$ \\
\hline
$\Delta$: (4)-(1) & 13.06 & 37.66 & -0.93 & 8.53 & 3.89 & 14.96 & 2.63 & 10.81 & 8.42 & 1.34 \\
$\Delta$: (5)-(1) & 13.54 & 37.58 & -1.41 & 8.64 & 3.89 & 15.63 & 2.33 & 10.50 & 8.01 & 1.39 \\
$\Delta$: (4)-(2) & 0.11 & -0.02 & 1.04 & 0.29 & 0.76 & -0.10 & 0.49 & 0.93 & 0.61 & 0.24 \\
$\Delta$: (5)-(2) & 0.60 & -0.10 & 0.55 & 0.40 & 0.75 & 0.57 & 0.20 & 0.62 & 0.20 & 0.29 \\
$\Delta$: (4)-(3) & 0.91 & 0.88 & 1.53 & 1.42 & 1.00 & 0.80 & 1.16 & 1.24 & 0.35 & 0.08 \\
$\Delta$: (5)-(3) & 1.40 & 0.79 & 1.04 & 1.53 & 0.99 & 1.47 & 0.86 & 0.93 & -0.06 & 0.13 \\
 \hline\hline
 & hr & hu & hy & kk & ko & lt & ms & no & pl & pt \\
 \hline
(1): {\bf\textsc{Teacher}}  & 50.53 & 52.49 & 21.55 & 22.82 & 26.88 & 45.35 & 24.09 & 62.76 & 56.53 & 51.77 \\
(2): {\bf\textsc{Baseline}}  & 60.19 & 62.75 & 32.32 & 35.85 & 35.56 & 52.31 & 24.76 & 67.38 & 69.31 & 52.10 \\
(3): {\bf\textsc{Emission}}  & 59.79 & 61.37 & 30.69 & 31.63 & 35.26 & 51.95 & 25.07 & 67.49 & 69.07 & 52.30 \\
(4): {\bf\textsc{Posterior}}  & 61.10 & $\mathbf{63.34}$ & 32.80 & $\mathbf{37.38}$ & 36.19 & 52.75 & $\mathbf{25.42}$ & $\mathbf{68.58}$ & $\mathbf{70.27}$ & 53.51 \\
(5): {\bf\textsc{Posterior+Top-K}}  & 60.58 & 63.21 & 32.57 & 34.10 & 36.70 & $\mathbf{52.83}$ & 25.14 & 67.51 & 69.90 & $\mathbf{53.53}$ \\
\hline
$\Delta$: (4)-(1) & 10.57 & 10.85 & 11.25 & 14.56 & 9.31 & 7.40 & 1.33 & 5.82 & 13.74 & 1.74 \\
$\Delta$: (5)-(1) & 10.05 & 10.72 & 11.02 & 11.28 & 9.82 & 7.48 & 1.05 & 4.75 & 13.37 & 1.76 \\
$\Delta$: (4)-(2) & 0.91 & 0.60 & 0.49 & 1.53 & 0.63 & 0.44 & 0.66 & 1.20 & 0.96 & 1.42 \\
$\Delta$: (5)-(2) & 0.40 & 0.46 & 0.25 & -1.75 & 1.14 & 0.53 & 0.38 & 0.13 & 0.58 & 1.44 \\
$\Delta$: (4)-(3) & 1.31 & 1.97 & 2.12 & 5.75 & 0.93 & 0.80 & 0.35 & 1.09 & 1.20 & 1.22 \\
$\Delta$: (5)-(3) & 0.79 & 1.83 & 1.88 & 2.47 & 1.44 & 0.88 & 0.07 & 0.03 & 0.83 & 1.23 \\
 \hline\hline
 & ro & ru & sk & sv & tr & uk & vi & zh & Avg. &  \\
 \hline
(1): {\bf\textsc{Teacher}}  & 34.96 & 21.91 & 52.84 & $\mathbf{70.44}$ & 45.98 & 25.04 & 30.05 & 3.40 & 41.85 &  \\
(2): {\bf\textsc{Baseline}}  & 36.46 & 28.68 & 60.44 & 68.91 & 57.14 & $\mathbf{49.19}$ & 33.38 & 28.94 & 50.86 &  \\
(3): {\bf\textsc{Emission}}  & 36.20 & 28.63 & 60.08 & 69.48 & 56.29 & 46.23 & 33.27 & 27.25 & 50.19 &  \\
(4): {\bf\textsc{Posterior}}  & $\mathbf{37.06}$ & $\mathbf{29.07}$ & $\mathbf{61.09}$ & 68.23 & $\mathbf{57.88}$ & 48.76 & $\mathbf{33.64}$ & $\mathbf{30.15}$ & $\mathbf{51.43}$ &  \\
(5): {\bf\textsc{Posterior+Top-K}}  & 36.33 & 29.05 & 60.78 & 69.30 & 57.68 & 46.82 & 33.28 & 30.04 & 51.14 &  \\
\hline
$\Delta$: (4)-(1) & 2.10 & 7.16 & 8.25 & -2.21 & 11.90 & 23.72 & 3.59 & 26.75 & 9.58 &  \\
$\Delta$: (5)-(1) & 1.37 & 7.14 & 7.94 & -1.14 & 11.70 & 21.78 & 3.23 & 26.64 & 9.29 &  \\
$\Delta$: (4)-(2) & 0.60 & 0.40 & 0.65 & -0.68 & 0.74 & -0.44 & 0.26 & 1.21 & 0.57 &  \\
$\Delta$: (5)-(2) & -0.13 & 0.37 & 0.34 & 0.39 & 0.54 & -2.38 & -0.10 & 1.11 & 0.28 &  \\
$\Delta$: (4)-(3) & 0.86 & 0.45 & 1.01 & -1.25 & 1.59 & 2.53 & 0.37 & 2.90 & 1.23 &  \\
$\Delta$: (5)-(3) & 0.13 & 0.42 & 0.70 & -0.18 & 1.39 & 0.58 & 0.02 & 2.79 & 0.94 & \\
\hline\hline
\end{tabular}
\caption{F1 scores of zero-shot transfer on the WikiAnn NER datasets. $\Delta$ represents the difference of F1 score.}
\label{tab:zs_ner}
\end{table*}

\begin{table*}[t]
\centering
\small
\begin{tabular}{l|ccccccccc}
\hline
\hline
& ar & bg & ca & cs & da & de & es & eu & fi  \\
\hline
(1): {\bf\textsc{Teacher}}  & 47.85 & 48.24 & 80.04 & 51.62 & 53.79 & 44.35 & 81.03 & 44.29 & 51.50 \\
(2): {\bf\textsc{Baseline}}  & 80.82 & 88.59 & 89.95 & 87.55 & 88.35 & 87.70 & 91.32 & 69.62 & 80.06 \\
(3): {\bf\textsc{Emission}}  & 80.85 & $\mathbf{88.62}$ & 90.00 & $\mathbf{87.56}$ & 88.47 & $\mathbf{87.89}$ & 91.27 & 69.68 & 80.10 \\
(4): {\bf\textsc{Posterior}}  & $\mathbf{80.95}$ & 88.26 & 89.77 & 87.50 & $\mathbf{88.68}$ & 87.79 & $\mathbf{91.48}$ & 70.03 & $\mathbf{80.52}$ \\
(5): {\bf\textsc{Posterior+Top-K}}  & 80.77 & 88.30 & 89.77 & 87.46 & 88.58 & 87.84 & 91.29 & $\mathbf{70.17}$ & 80.38 \\
\hline
$\Delta$: (4)-(1) & 33.10 & 40.02 & 9.73 & 35.88 & 34.89 & 43.44 & 10.45 & 25.74 & 29.02 \\
$\Delta$: (5)-(1) & 32.92 & 40.06 & 9.73 & 35.84 & 34.79 & 43.49 & 10.26 & 25.88 & 28.88 \\
$\Delta$: (4)-(2) & 0.12 & -0.33 & -0.18 & -0.05 & 0.33 & 0.09 & 0.15 & 0.41 & 0.47 \\
$\Delta$: (5)-(2) & -0.05 & -0.30 & -0.18 & -0.09 & 0.23 & 0.14 & -0.03 & 0.55 & 0.32 \\
$\Delta$: (4)-(3) & 0.09 & -0.36 & -0.24 & -0.06 & 0.21 & -0.10 & 0.20 & 0.34 & 0.42 \\
$\Delta$: (5)-(3) & -0.08 & -0.33 & -0.23 & -0.10 & 0.11 & -0.05 & 0.02 & 0.49 & 0.28 \\
 \hline\hline
 & hi & hr & it & ko & nl & no & pl & pt & ro \\
 \hline
(1): {\bf\textsc{Teacher}}  & 33.09 & 69.40 & 79.33 & 37.90 & 40.02 & 50.86 & 48.68 & 77.66 & 70.45 \\
(2): {\bf\textsc{Baseline}}  & 76.41 & 88.28 & 93.66 & 58.47 & 87.30 & 88.84 & 85.26 & 93.38 & 86.20 \\
(3): {\bf\textsc{Emission}}  & 76.15 & 88.17 & 93.74 & 58.65 & $\mathbf{87.32}$ & $\mathbf{88.94}$ & 85.27 & $\mathbf{93.49}$ & 86.15 \\
(4): {\bf\textsc{Posterior}}  & $\mathbf{76.64}$ & $\mathbf{88.46}$ & 93.70 & $\mathbf{59.09}$ & 87.19 & 88.91 & 85.31 & 93.42 & 86.33 \\
(5): {\bf\textsc{Posterior+Top-K}}  & 76.44 & 88.34 & $\mathbf{93.83}$ & 58.85 & 87.20 & 88.83 & 85.60 & 93.15 & $\mathbf{86.57}$ \\
\hline
$\Delta$: (4)-(1) & 43.55 & 19.06 & 14.37 & 21.19 & 47.17 & 38.05 & 36.63 & 15.76 & 15.88 \\
$\Delta$: (5)-(1) & 43.35 & 18.94 & 14.50 & 20.95 & 47.18 & 37.97 & 36.92 & 15.49 & 16.12 \\
$\Delta$: (4)-(2) & 0.23 & 0.18 & 0.03 & 0.62 & -0.11 & 0.07 & 0.05 & 0.03 & 0.13 \\
$\Delta$: (5)-(2) & 0.03 & 0.06 & 0.17 & 0.38 & -0.10 & 0.00 & 0.34 & -0.23 & 0.36 \\
$\Delta$: (4)-(3) & 0.50 & 0.29 & -0.05 & 0.45 & -0.13 & -0.03 & 0.04 & -0.07 & 0.18 \\
$\Delta$: (5)-(3) & 0.30 & 0.18 & 0.09 & 0.21 & -0.11 & -0.10 & 0.33 & -0.34 & 0.41 \\
 \hline\hline
 & ru & sk & sr & sv & tr & zh & Avg. &  &  \\
 \hline
(1): {\bf\textsc{Teacher}}  & 50.81 & 56.09 & 70.04 & 50.63 & 54.93 & 51.55 & 56.01 &  &  \\
(2): {\bf\textsc{Baseline}}  & 88.15 & 87.67 & 89.70 & 89.73 & 71.49 & 70.24 & 84.11 &  &  \\
(3): {\bf\textsc{Emission}}  & 88.10 & 87.73 & 89.60 & 89.91 & 71.68 & 70.72 & 84.17 &  &  \\
(4): {\bf\textsc{Posterior}}  & $\mathbf{88.22}$ & 87.83 & $\mathbf{89.95}$ & $\mathbf{89.96}$ & 71.93 & $\mathbf{70.93}$ & $\mathbf{84.28}$ &  &  \\
(5): {\bf\textsc{Posterior+Top-K}}  & 88.10 & $\mathbf{87.84}$ & 89.92 & 89.69 & $\mathbf{71.99}$ & 70.74 & 84.24 &  &  \\
\hline
$\Delta$: (4)-(1) & 37.41 & 31.74 & 19.91 & 39.33 & 17.00 & 19.38 & 28.28 &  &  \\
$\Delta$: (5)-(1) & 37.29 & 31.75 & 19.88 & 39.06 & 17.06 & 19.19 & 28.23 &  &  \\
$\Delta$: (4)-(2) & 0.07 & 0.16 & 0.25 & 0.23 & 0.44 & 0.69 & 0.17 &  &  \\
$\Delta$: (5)-(2) & -0.05 & 0.18 & 0.22 & -0.04 & 0.50 & 0.50 & 0.12 &  &  \\
$\Delta$: (4)-(3) & 0.12 & 0.10 & 0.35 & 0.05 & 0.24 & 0.21 & 0.12 &  &  \\
$\Delta$: (5)-(3) & 0.00 & 0.11 & 0.32 & -0.21 & 0.31 & 0.02 & 0.07 &  & \\
\hline\hline
\end{tabular}
\caption{F1 scores of zero-shot transfer on the UD POS tagging datasets. $\Delta$ represents the difference of F1 score.}
\label{tab:zs_pos}
\end{table*}

\end{document}